# Semantic Cells
# Evolutional Process to Acquire Sense Diversity of Items

Yukio Ohsawa,    Dingming Xue,    Kaira Sekiguchi
The University of Tokyo
info@panda.sys.t.u-tokyo.ac.jp

**Abstract.** Previous models for learning the semantic vectors of items and their groups, such as words, sentences, nodes, and graphs, using distributed representation have been based on the assumption that the basic sense of an item corresponds to one vector composed of dimensions corresponding to hidden contexts in the target real world, from which multiple senses of the item are obtained by conforming to lexical databases or adapting to the context. However, there may be multiple senses of an item, which are hardly assimilated and change or evolve dynamically following the contextual shift even within a document or a restricted period. This is a process similar to the evolution or adaptation of a living entity with/to environmental shifts. Setting the scope of disambiguation of items for sensemaking, the author presents a method in which a word or item in the data embraces multiple semantic vectors that evolve via interaction with others, similar to a cell embracing chromosomes crossing over with each other. We obtained two preliminary results: (1) the role of a word that evolves to acquire the largest or lower-middle variance of semantic vectors tends to be explainable by the author of the text; (2) the epicenters of earthquakes that acquire larger variance via crossover, corresponding to the interaction with diverse areas of land crust, are likely to correspond to the epicenters of forthcoming large earthquakes.

## 1   Introduction

Semantic vectors were invented in the 1960s, and have been applied to natural language analysis and large language models [Camacho-Collados and Pilevar 2018]. The approaches mapping the semantics of words into low-dimensional vectors, such as Vector Space Model [Schütze et al 1993, Reisinger et al 2010] or Word Embedding [Y. Bengio et al 2003], have proven their significance in capturing syntactic and semantic information in multiple Natural Language Processing (NLP) downstream tasks, e.g., machine translation [D. Bahdanau et al 2015], natural language inference [Conneau et al 2018, Talman et al 2018], semantic analysis [Landauer et al 1998], text classification [Kim 2006] and so on.

Research on distributed semantic representations has gone through several developmental stages, with the core theories transitioning from embedding the words' index in a corpus [Y.Li et al 2018] to bi-directional language modeling considering the semantics of context words [Devlin et al 2019]. A simple and conventional method of representing a text word as a one-hot binary vector was formed as the basis of NLP and can fulfill the needs of NLP tasks, such as document classification [Li et al 2018]. In this method, each word is represented as a binary vector of vocabulary size. The representation is rather simple, and its limitation is that it does not consider the semantic relationships between adjacent words or context insensitivity [Zhang et al 2010]. With the maturity of the distribution hypothesis, posits that contextually similar words have similar semantics, making word embeddings different in different contexts, and can address disambiguation to some extent [Wang et al 2020]. These word representation methods are mainly divided into three types: matrix-factorization-based representation [Collobert et al 2008, Lebret et al 2013]，sliding-window-based methods [Mikolov et al 2013ab], and neural network-based methods [Xu et al 2000]. The embeddings based on the distribution hypothesis were determined after training, and the word embeddings were static.

Using semantic vectors which are the embeddings learned from a large collection of text, the arithmetic operation came to be applied as a tool for reasoning the sense of a composite words such as "sister" as a sum of if compounds "brother" and "woman" because "sister" is a composition of two contexts (*sibling* and *woman as a gender category*). However, because the word's sense "sister" is represented by a single semantic vector, the word "sister" meaning "nun" is not interpret as it should be. Furthermore, the sense of a word changes even within a domain according to the contextual shifts although the change may be less radical e.g. "sister" does not turn to mean a nun but changes from one's sister as a member of the same family living in the same culture into a member of another family in a different culture after marriage. Thus, the change in sense *may* occur as a continuous shift in context rather than a discontinuous jump across domains.

Word Sense Disambiguation (WSD) is considered a long-term task in NLP. According to Navigli [Navigli et al 2013], the WSD task involves selecting the intended sense from a predefined set of senses for that word defined by a sense inventory. Here, we consider how WSD can be realized to foster sensemaking in data [Dervin 1992], which is a process toward collection, representation, and organization of information decision-making in real tasks [Russell et al. 1993, Weick 1993]. This is supposed to be initiated by recognizing the inadequacy of the current understanding of events to explain their meaning [Klein et al. 2006b]. The ambiguity of the sense of an event, item, or word disturbs sensemaking, which is a problem that humans may solve if the available information is of a moderate size, but is difficult to solve if it is large or too small. Hence, machine-aided WSD is required.

The sheer method for assigning a semantic vector to each word can represent multiple senses of a word as long as only a single vector corresponds to a word. If the vector for a polysemous word, such as a sister, is given by the average of all its senses, the essential multiple senses are lost. Therefore, methods have been developed to vectorize word senses rather than words. According to [Camacho-Collados and Pilevar 2018], there are mainly two types of approaches: unsupervised and knowledge-based. For example, methods have been proposed to choose a suitable semantic vector for each word by automatically estimating the word-sense inventory during training [Cheng and Kartsaklis 2015, Li and Jurafsky 2015a, Neelakantan et al 2014, Tian 2014]. For example, [Neelakantan et al 2014] estimated the sense of a word from its context during training, and adaptively selected and revised a cluster of embeddings from which to choose the semantic vector. On the other hand, some studies utilized lexical databases containing semantic networks [Faruqui et al 2015a, Jauhar et al 2015, Rothe and Schütze 2015], among which [Jauhar et al 2015] considered that the semantic vectors are learnt under constraints defined by the structure of the semantic network and allowed the word sense disambiguation task to be performed with high accuracy. These methods acquire a distributed representation of each item while conforming to existing lexical databases.

To generate the distributed representation of a word in each sentence or partial word series in a text, a single word is associated with various vectors, depending on the thickness of the context in which it appears. For example, as seen in [Maehara and Takenaka 2023], sense in a sentence can be vectorially represented by acting on the semantic vector of a word, with vectors representing the context before and after it. Amrami [Amrami et al 2018] proposed a method using the LSTM language model to predict probable substitutes for target words and induce senses by clustering these resulting substitute vectors. Doc2vec [Le and Mikolov 2014], extending word2vec from learning embeddings of words to those of word sequences on the paragraph vector-distributed bag of words (PV-DBOW) (interchangeably referred to as doc2vec skip gram), enabled the learning of vector representations of arbitrary-length word sequences such as sentences, paragraphs, or a document. However, this approach partially loses the strength of using lexical databases because it is not based on the diversity of the original senses of a word itself, as the effect of context is reflected in the pre-learned vectors for words.

In contrast, since 2018, the mainstream approach has used context-embedded distributed representations in BERT and other approaches, where the polysemy of words and diversity of expressions are pre-learned in the transformer model. Following LSTM-based ELMo, [Peters et al 2018], transformer-based GPT, BERT [Devlin et al 2019] and other models have been established for generating dynamic, deep contextualized embeddings. Through deep language modeling, they can grasp the semantics of words with contextualized meanings more accurately. Then, [Wiedemann et al 2019] argued that a pre-trained BERT model could place distinct senses of a word corresponding to distinct places in the embedding space. As for transformer-based langue models, Zhou [Zhou et al 2019] proposed an end-to-end BERT-based lexical substitution approach which can validate substitute candidates without using any annotated data or manually curated resources. This method tends to consider the substitution's influence on the global context of the sentence and successfully avoids the issue of overlooking the candidates that are not synonyms of the target words while using lexical resources. However, the problem is that training neural networks such as LSTM and transformers require huge amounts of data, making it difficult to apply these techniques to phenomena that are rarely observed and have small datasets.

The aim of the present study is to acquire the semantic vector for each word that may change from/to sentences, as in the above example of sister, but is actually linked to one of several fixed candidate senses depending on the context. In other words, we aim to first obtain the semantic vectors of multiple specific candidates and choose the most suitable one that fits the context, to which a certain fine-tuning computation is used.

As various datasets have been studied targeting real-world domains on their analogy with text and the senses of items in each domain have been put into vectors, the proposed method aims to divert to various domains. Co-occurrence based analyses of earthquakes [Ohsawa 2002, Fukui and Inaba 2014] and vectors for n-grams in biological sequences (e.g., DNA, RNA, and proteins) have been proposed [Asgari and Mohammad 2015], and commercial items in the market have also been vectorized [Barkan and Koenigstein 2016, Cao et al 22]. In the extension of these studies, a basic algorithm for WSD can be diverted to the vectorization of various items to understand the sense (the ways to use) of daily commodities and the effects of disasters. This study realizes a method to deal with multiple senses with a mechanism that can be applied to datasets of rare events, such as large earthquakes.

In this study, as described in Section 2, we developed a novel method to approach a WSD that satisfies the aforementioned aim. Each sense of a word is represented by a vector and multiple vectors correspond to a set of senses of the word. The approach here to WSD is to foster the evolution of semantic vectors forth and back between 'universality of the word sense and 'adaptability to the context' by introducing a genetic computational framework. Each vector is regarded as a chromosome, and the word represents a cell-embracing chromosome. Similar to the crossover of individuals living together, words exchange parts of their vectors if they appear in the same sentence. Chromosomes that participate in the crossover interaction are selected from cells according to their distance from each other. After one round of reading the target text, each word acquires its own diversity of senses as a diversity of semantic vectors in its cell. As a word representing a novel idea can be proposed by connecting pieces of knowledge from multiple domains, we expect that this diversity of senses of a word appearing multiple times in a document may lead to the appearance of a novel concept. This relevance of a word to multiple domains is obtained via crossover and results in the diversity of chromosomes corresponding to the item (word). As described in Section 3, this effect is observed in the correspondence of high-role words, whose semantic roles in the text can be explained without reviewing the target text, with words of high or lower-middle semantic diversity acquired via the crossover process. That is, high-diversity words are expected to contribute to creativity, whereas lower-middle-diversity words correspond to the essential basic concepts used for creation. As extended in Section 3, to a dataset where the items are *not* words in text but earthquakes - events in nature. We expect to extend Semantic Cells to social applications, including items in the market, for which analogy with text has been discovered so far, and further to various real-world problems.

## 2 Semantic Cells

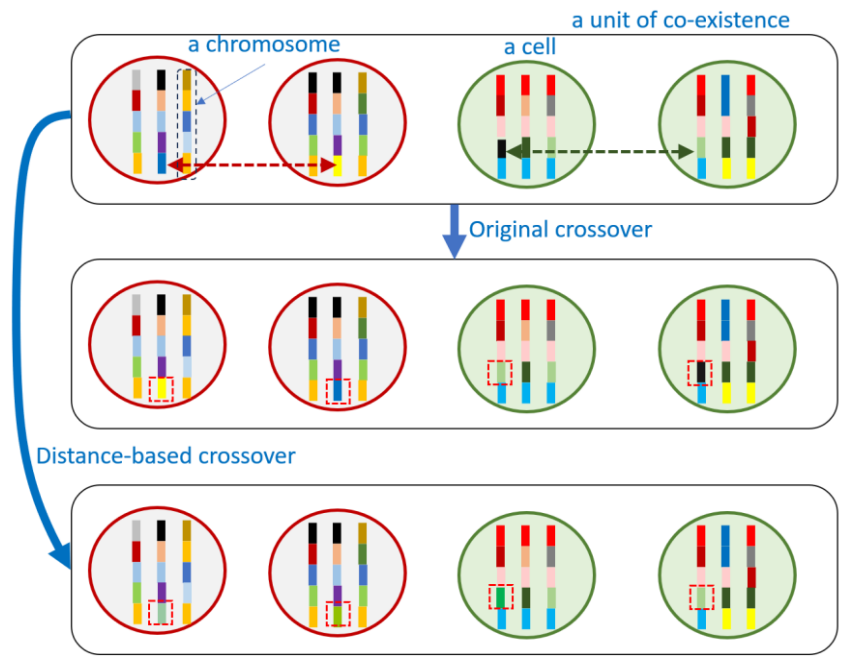

**Figure 1.** The crossover between chromosomes in cells, each of which corresponds to an item (word, item, event, etc.). Multiple chromosomes in a cell correspond to the multiple senses of an item, among those within close relationships (similarities) execute a crossover. The crossover is simply realized in this study by reducing the distance between two chromosomes by changing the values of elements of the vectors corresponding to each genetic element.

As shown in Figure 1, a Semantic Cell $C_w$ is assigned to item $w$ in the itemset $W$. Here, an item refers to a word if the dataset is text, a commercial item if POS data, an earthquake or its focal point (or epicenter) if it is historical seismic data. $C_w$ includes $g$ chromosomes, which are vectors representing the $g$ senses of an item. A sense may correspond to the meaning if the item is a word, customer's utility if the item is a commercial item sold in the market, or a source of the force if the item is an earthquake.

If item $w$ occurs with other items in the $j$-th unit of co-existence, for example, a sentence in text or a basket in the POS data of a supermarket, the chromosomes go through the process of crossover which has been used in genetic algorithms [Goldberg and Lingle 1985, Goldberg 1989, Eiben and Smit 2015ab]. In contrast to existing methods of genetic algorithms for the crossover operation, where parts of a chromosome are exchanged with others, we moved selected chromosomes close to others in cells in the same coexistence unit. The selection in the preliminary version of the semantic cell is executed on the distance (the closer it is, the more likely it is to be selected) with other chromosomes in the same coexistence unit. The number of crossovers equals to the number of coexisting units in the dataset is called one round, and $R$ rounds are executed in a crossover operation. The final vectors, that is, chromosomes, as a result of the crossover process, represent the diversity within a cell.

The evolution procedure of semantic cells can be described as follows, where $d$ represents the dimension of each vector, which is the length of, that is, the number of genes in, each chromosome. $g$ is the number of chromosomes (i.e., vectors) in each cell (i.e., item). $R$, the number of rounds in which the crossover is iterated, which was set to one in this study as a preliminary experiment. In each crossover operation, $Ch_S$ is used represent the vector corresponding to the co-existence unit $S$. The crossover in Figure 1 is executed by reducing the distance between the chromosome selected from each cell and all other chromosomes in the belonging co-existence unit, where the selection is simply distance-based; that is, the chromosome in a cell with the shortest distance to $Ch_S$ is selected. Although $Ch_S$ can be computing the vector of the co-existence unit using the method for doc2vec [Le and Mikolov 2014], here we used the sheer average of all the vectors in all the cells in in unit $S$ for simplicity. $\alpha$ here represents the influence of a unit on the selected chromosome of an item during the crossover operation. Although $\alpha$ may be replaced by $\alpha/r^2$, where $r$ is the distance above, considering the analogy with gravity or Coulomb's force between two particles, we used only $\alpha$ to accelerate the crossover of items that have been located far from each other in the distance-based crossover in Line 13.

In line 4, we applied word2vec to the Gensim library [Řehůřek 2022] to obtain the initial chromosomes, that is, semantic vectors of words, where $d$ was set to 50 and $g$ to 5, which means that we set the preliminary experimental condition to a small size and the upper number of senses of each word to as small as five. These settings may be simpler than real natural language or standard values of $d$ because this is a preliminary experiment. However, it should be noted that a larger dimension does not necessarily result in a more accurate analysis [Melamud et al 2016].

1: **Input:**
2:   Dataset $D$ including co-existence units $\{S|\ S \subset D\}$
3:   Set of items $\{w|$ each item in unit $S$ including $|S|$ items$\}$
4:   The $j$-th chromosome $\text{Ch}_{wj}$, of item $w$ including $g$ chromosomes
5:   Each gene: $\text{chr}_{wjk}|\ 0 < j \leq g,\ 0 < k \leq d, w \in W$
6: **Output:**
7:   Evolved chromosomes $\text{Ch}_{wj}$ for each $w \in W\ and\ j \in g$
8: **Crossover iteration:**
9:    for each $r$ in $[1:R]$
10:       for each $S \subset D$
11:          for each $w \in S$
12:             $\text{Ch}_{wj_0}$ = the $j_0$th chromosome in $w$, of the least distance in $w$ from $\text{Ch}_S$
13:             $\text{Ch}_{wj_0} = (1-\alpha)\text{Ch}_{wj_0} + \alpha\ \text{Ch}_S$
14:          end for
15:       end for
16:    end for

## 3   Preliminary Experiments

### 3.1 Applying semantic cells to text

Here, we first applied Semantic Cells to small text in natural language. To understand its qualitative performance rather than jumping to statistical evaluation by applying it to big data, we applied the method to four papers and one Call for Papers written with collaborators [Ohsawa et al 2022, 2023, 2024ab, IEEE Bigdata 2024]. We chose these texts to evaluate the meaningfulness of the diversity of the senses obtained using Semantic Cells. That is, the diversity of the senses of a word means the diversity of the contexts in which the word has been used, as in the example of "sister." A word is used in various contexts when it is related to various concepts in the text; for example, in the preface, main content, exemplifications, and/or conclusions, which implies that the word represents an essential point of the author of the target document. On the other hand, a word that is essential in a specific knowledge domain and serves as a basis for asserting a new point. Thus, observing the bipolar distribution of diversity among essential words is more essential than sheer statistic evaluation of accuracy, that is, words of larger diversity and of lower diversity, are both expected to play essential roles in constructing the architecture of a document that the author of the target text writes.

To observing the bipolar distribution of diversity among essential words, we computed the semantic diversity of each word $w$ as the sum of the variances of the $d$-dimensional values of the $g$ chromosomes as in Eq.(1):

$$div(w) = \sum_{k}^{d} var_{j \in [1:g]}[\text{chr}_{wjk}] \qquad (1)$$

Then, the author scored 1 if the role of the word $w$ of the $i$-th ranked $div(w)$ in his own thought could be explained; otherwise 0. Then, the smoothed score of explainability for the $i$-th rank was obtained as the average of the $i$-th through $i$+100 th ranked words. The result is shown in Figure 2. In Figure 2, the horizontal axis shows the smoothed score of explainability for the corresponding rank of the semantic diversity. As shown in this figure, the highest and the middle-range semantic diversities correspond to words of high explainability, which is here regarded as an index of importance of words. Thus, our hypothesis above is here validated for this preliminary stage. The words that appear on the two hills in Figure 2 are listed in Table 1. The words on the top (high-rank) hill represent ideas with ambiguities, which were used in multiple contexts. Let us present the following examples.

- "FC": a *Feature Concept*, which stands for an abstraction of knowledge or concept that is expected to be acquired from a dataset or combined datasets. This is an ambiguous word because an FC may refer to an image, illustration, or text, depending on the target domain and context of data utilization.
- "DL" and "Data Leaves": representations of the knowledge which is expected to be discovered from a certain given dataset. This is also an ambiguous word in a sense similar to FC. Both FC and DL appear in multiple contexts corresponding to their application domains in the target text in this experiment [Ohsawa 2024a].
- "Physical": means things and matters related to bodies, food, and well-being, related to (1) human body or (2) the physical world in comparison with the cyber world.
- "Communities": a concept used in the target text, meaning both (1) groups of people sharing interests and (2) regions from which people are originated.
- "Well-being": has at least five meanings which are financial wellbeing, physical wellbeing, career wellbeing, community wellbeing, and social wellbeing as surveyed and separately dealt in the target text [Ohsawa 2024b]

In contrast, the words on the right side of Table 1 were used to refer to essential but specific senses as follows:

- "smartphones" "Collect_connect_DLs_to_G," "Create_FC_graph": tools used for collecting and analyzing data, in specific technical parts of the target text.
- "COVID-19," "Gonorrhoea," "Chlamydia," "Influenza": diseases as specific targets of the authors' analysis. Especially, "Gonorrhoea," "Chlamydia," and "Influenza" appeared in only two paragraphs in the target text.
- "Haseko" and "Suumo": referred to as specific providers of information used in a part of studies represented in the target text.
- "Entropy": starting from the capital letter E, different from "entropy" in the left half of Table 1, used only as a word in a specific method.

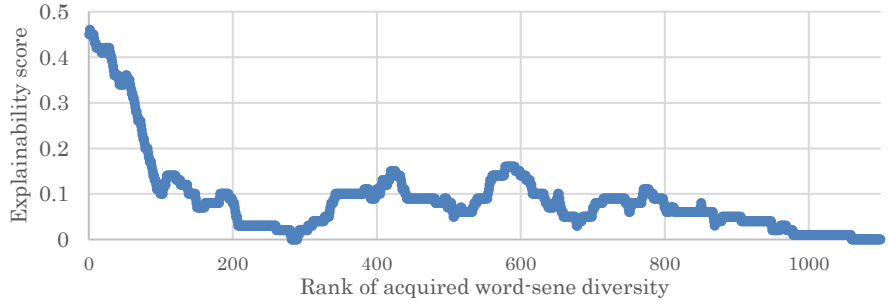

**Figure 2**. A result of a case where the Semantic Cell was applied to text. Words with the largest diversity of senses as a result of crossover tend to be the most explainable, and the middle-lower (350th - 800th) has a lower peak corresponding to specific and essential bases of the author's points.

**Table 1.** Words selected from the top hill (the left half here, from the 1st to 200th in Figure 2) and the middle-lower (the right half here, from the 350st to 800th in Figure 2).

| rank | word | variance | rank | word | variance |
|---|---|---|---|---|---|
| 2 | FC | 0.39 | 391 | COVID-19 | 0.06 |
| 6 | DL | 0.27 | 403 | smartphones | 0.06 |
| 7 | regions | 0.27 | 481 | inter-community | 0.05 |
| 8 | well-being | 0.27 | 497 | moving-direction | 0.05 |
| 10 | individuals | 0.26 | 498 | intercommunity | 0.05 |
| 17 | crowds | 0.25 | 504 | Multiscale | 0.05 |
| 28 | physical | 0.23 | 505 | entropies | 0.05 |
| 35 | communities | 0.20 | 506 | geographical | 0.05 |
| 38 | crowd | 0.20 | 518 | Collect_connect_DLs_to_G | 0.05 |
| 39 | inter-crowd | 0.20 | 615 | Haseko | 0.05 |
| 57 | Data | 0.18 | 616 | Suumo | 0.05 |
| 93 | Leaves | 0.15 | 654 | Gonorrhoea | 0.04 |
| 98 | diversity | 0.14 | 655 | Chlamydia | 0.04 |
| 99 | DLs | 0.14 | 656 | Influenza | 0.04 |
| 100 | people | 0.14 | 659 | Create_FC_graph | 0.04 |
| 121 | entropy | 0.13 | 677 | Entropy | 0.04 |

**3.2 Applying semantic cells to earthquakes**

Second, we applied Semantic Cells to the earthquakes. The aim of this study is to highlight the precursory events of forthcoming disasters, which correspond to sensemaking from earthquake data. The risk of earthquakes in a region has been estimated using various methods based on the distribution of earthquakes and spatiotemporal distances. For established reviews of seismic precursors, see [Wyss and Habermann 1988, Mignan 2011, Mignan 2014] where the debates around the prognostic value of precursors as well as the different schools of thought is described.

With the development of computing algorithms, purely data-driven approaches have been developed for earthquake prediction. For example, the eigenvectors and corresponding eigenvalues of the matrix representing pairwise co-occurrences of earthquakes have been used to predict the probability of earthquake occurrence in clusters of regions [Fukui 2014]. However, the precursors of large earthquakes are difficult to capture using this approach owing to their complex and unknown latent dynamics. The epidemic-type aftershock sequence (ETAS) also shows good performance in estimating the risk in regions where earthquakes frequently occur [Ogata and Zhang 2006] and has been extended in various directions, such as coping with biases of ETAS [Grim et al 2022]. However, some earthquakes beyond the reach of these models show significant exposure of energy, particularly in regions where the frequency of earthquakes is low. The interaction and variety of earthquake activities across a wide region have also been considered in pattern informatics [Nanjo 2006ab] where an earthquake is assumed to be a multibody phenomenon ruled by latent dynamics of lithosphere. The pattern informatics method has also been improved by combining regions to select for parameter normalization [Tian et al 2024]. In this study, we borrow the basic idea of a cluster-based analysis of data from an earthquake catalog [Ohsawa 2018] where the co-occurrence of earthquakes and the distances among epicenters were used to extract the iinteraction among multiple clusters to highlight regions of near-future risks.

Here, each individual cell corresponds to one mesh region of width (0.5, 0.5) in latitude and longitude on the land crust, which may include none or multiple epicenter(s) appearing in the earthquake catalog provided by the Japan Meteorological Agency (JMA), including the time, latitude, longitude, magnitude for each earthquake: open access at (http://www.data.jma.go.jp/svd/eqev/data/bulletin/hypo.html), the data site of JMA. The conditions for modifications and uses of this map are available at http://ww.jma.go.jp/jma/en/copyight.html. Thus, each event of an earthquake is considered as the appearance of a cell in a co-existence unit, which is a set of earthquakes that occurred within a certain period, that is, ten consecutive earthquakes. The initial values of the genes in the chromosomes, as in Line 4 in the pseudocode above, are given as the 2-dimensional geometrical coordinates of the epicenters. Then, in the crossover operation, the chromosome in each cell, that is, an epicentral region, which is closest to the virtual chromosome averaging all the chromosome vectors in all the cells in the co-existence unit, moves even closer to the average chromosome as in Line 13. As in the results in Figure 3, the colors of the cells are shown on the ranks of their semantic diversities and crossover operations: red (orange) as the original positions of cells ranked in the top 15 (50) diversity of chromosomes, and blue as the 250 chromosomes which were obtained as a result of the crossover for the five chromosomes of the 50 cells in orange.

Based on the assumption that a region interacting with diverse surrounding regions is associated with a high risk of earthquakes, that is an idea we borrowed from [Ohsawa 2018], the red and orange regions are regarded as candidates for high-risk regions. Comparing these with the red crosses showing the epicenters of real earthquakes of large (> M6) earthquakes within two years from the end of the period of the dataset (e.g. from Jan 2011 till Dec 2011 corresponding to the earthquake catalog for year 2010), we can summarize the features of Figure 3 as follows:

(i) In a wide view, the red dots correspond roughly to the epicentral regions of large (> M6) earthquakes which attacked the archipelago of Japan. Furthermore, we found a correspondence between the local events and the locations of new red dots. For example, the epicenters of East Japan (M9.0, 2011), Noto (M7.2, 2024), and East Ehime (M6.6, 2024s) encountered earthquakes within two years after the appearance of red dots in the corresponding regions.
(ii) The blue dots (the 250 destinations of highest-diversity cells) are concentrated on the line from the northeast to the right-west, which moved to the east after M9.0 in 2011, and reshaped into straight lines in 2021 and 2022.

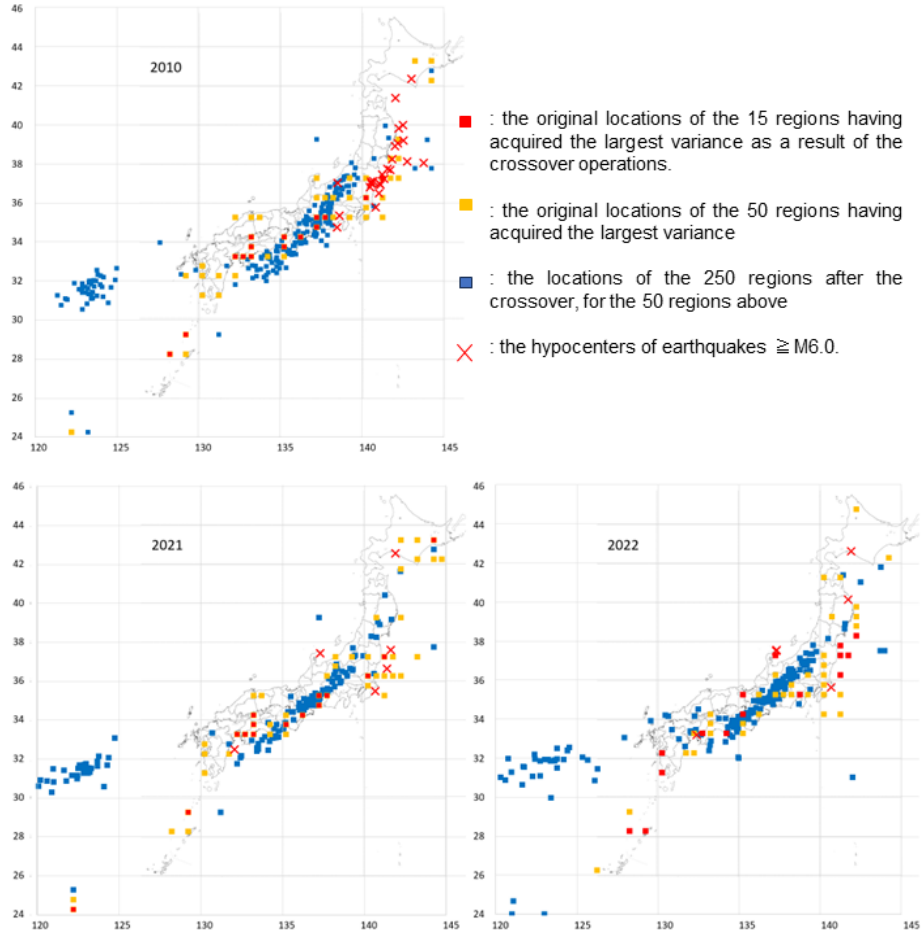

**Figure 3** The visualized results of semantic cells applied to seismic history in the open data provided by Japan Meteorological Agency (JMA); Red (orange) dots: the original locations of the 15 (50) regions of the highest diversity after the crossover process, blue dots: the 250 dispersed regions, red crosses: the hypocenters of earthquakes larger than or equal to M6.0.

## 4. Conclusions and Future Work

Here, we proposed semantic cells that initially addressed WSD for text in natural language. Based on the analogy of natural language and other domains, we extended this method to the selection of precursory events of earthquakes. For text applications, the requirement to apply to large volumes of data is currently an established trend.

However, from our viewpoint to regard sense-making as an important fruit of data utilization, the addressed question here is whether we can obtain information of which the meaning can be explained for the user. For this reason, we started with small texts within the reach of the authors' explanation, leaving our application to LLM to the future work.

We expect to extend semantic cells, with improvements and customizations, to social applications including items in the market and human movements in the real space, of which the analogy with text has been found and applied so far, and further to various real-world problems. The expected results are the discoveries of items of potential future popularity in marketing because such items are preferred in multiple hidden contexts, which potentially implies they attract people from a wide range of communities, occupations, and regions. In other words, the evolutionary process of semantic cells borrowing from a genetic system means not sheer learning from data but the evolution of senses.

## Acknowledgment

This study was supported by JST Grant JPMJPF2013 (ClimCore), Q-Leap JPMXS0118067246, JSPS Kakenhi 20K20482, 23H00503, and MEXT Initiative for Life Design Innovation.

## References

[Amrami et al 2018] Amrami, A., and Goldberg, Y., 2018. Word sense induction with neural biLM and symmetric patterns. *In Proc. Conf. on Empirical Methods in Natural Language Processing* (EMNLP), 4860–4867.

[Asgari and Mohammad 2015] Asgari, E., Mohammad, MRK., 2015, Continuous Distributed Representation of Biological Sequences for Deep Proteomics and Genomics. *PLOS One* 10 (11): e0141287. doi:10.1371/journal.pone.0141287.

[Barkan and Koenigstein 2016] Barkan, O., Koenigstein, N., 2016, Item2vec: neural item embedding for collaborative filtering. *In IEEE International Workshop on Machine Learning for Signal Processing* (MLSP) doi:10.1109/MLSP.2016.7738886.

[Bengio et al 2003] Bengio, Y., Ducharme, R., Vincent, P., Janvin, C, 2003, A neural probabilistic language model. Journal of Machine Learning Research, 3(6):1137–1155.

[Camacho-Collados and Pilevar 2018] Camacho-Collados, J. & Pilevar, M.T., 2018, From Word To Sense Embeddings: A Survey on Vector Representations of Meaning. *J. Artificial Intelligence Research*. 63. 743-788. doi:10.1613/jair.1.11259.

[Cao et al 2022] Cao, J., Cong, X., Liu, T., Wang, B., 2022, Item Similarity Mining for Multi-Market Recommendation. *In Proc. Int'l ACMSIGIR Conference on Research and Development in Information Retrieval* (SIGIR), doi:10.1145/3477495.3531839.

[Cheng and Kartsaklis 2015] Cheng, J. and Kartsaklis, D., 2015, Syntax-aware multisense word embeddings for deep compositional models of meaning, *Proc. Conference on Empirical Methods in Natural Language Processing* (EMNLP), 1531-154. doi:10.18653/v1/D15-1177.

[Collobert et al 2008] Collobert, R., and Weston, J., 2008, A unified architecture for natural language processing: deep neural networks with multitask learning. *In Proc. Int'l Conf. Machine learning*, 160–167.

[Conneau A et al 2018] Conneau A, Kiela D, Schwenk H, Barrault L, Bordes A, 2018, Supervised learning of universal sentence representations from natural language inference data. In: Conference EMNLP, 670-680, doi:10.18653/v1/D17-1070 .

[Bahdanau et al 2015] Bahdanau, D., Cho, K., Bengio, Y., 2015, Neural machine translation by jointly learning to align and translate. In *Int'l Conf. Learning Representations,*

[Dervin 1992] Dervin, B., 1992, From the mind's eye of the user: The sense-making qualitative-quantitative methodology. In Glazier, J. and Powell, R. R. *Qualitative research in information management*, 61-84. Englewood, CA

[Devlin et al 2019] Devlin, J., Chang, MW., Lee, K., Toutanova, K., 2019, BERT: Pre-training of Deep Bidirectional Transformers for Language Understanding. *In Proc. Conf. North American Chap. Assoc. for Comp. Linguistics: Human Language Technologies* 1, 4171–4186. doi: 10.18653/v1/N19-1423.

[Eibdoi:en and Smith 2015a] Eiben, AE., Smith, JE., 2015, Recombination for Permutation Representation, *Introduction to Evolutionary Computing. Natural Computing Series* (2nd ed.). Berlin, Heidelberg: Springer, 70–74. doi:10.1007/978-3-662-44874-8.

[Eiben and Smith 2015b] Eiben, AE., Smith, JE., 2015, Recombination Operators for Real-Valued Representation". *Introduction to Evolutionary Computing. Natural Computing Series* (2nd ed.). Berlin, Heidelberg: Springer, 65–67. doi:10.1007/978-3-662-44874-8.

[Faruqui et al 2015a] Faruqui, M., Dodge, J., Jauhar, S. K., Dyer, C., Hovy, E. and Smith, N. A., 2015, Retrofitting word vectors to semantic lexicons, P*roc. of NAACL-HLT,* 1606-1615. doi:10.3115/v1/N15-1184

[Fukui et al 2014] Fukui, K., Inaba, D., Numao, M., 2014, Discovering seismic interactions after the 2011 Tohoku earthquake by co-occurring cluster mining. *Inf. Media Technol* 9, 886–895. doi:10.11185/imt.9.886

[Grimm et al 2022] Grimm, C., Hainzl, S., Käser, M. *et al.* 2022, Solving three major biases of the ETAS model to improve forecasts of the 2019 Ridgecrest sequence. *Stoch Environ Res Risk Assess* 36, 2133–2152. doi:10.1007/s00477-022-02221-2

[Goldberg and Lingle 1985] Goldberg, D., Lingle, R., 1985, Grefenstette, John J. (ed.), Alleles, loci, and the traveling salesman problem, *Proc.t Int'l Conf. Genetic Algorithms and Their Applications* (ICGA), 154–159.

[Goldberg 1989] Goldberg, D., 1989, *Genetic Algorithms in Search, Optimization and Machine Learning. Reading*, MA: Addison-Wesley Professional. ISBN 978-0201157673.

[IEEE Bigdata 2024] Synergizing Mobility Data for Creating and Discovering Valuable Places, CFP for Special Session, IEEE Bigdata 2024 https://www.panda.sys.t.u-tokyo.ac.jp/nigiwaiSSS.html

[Jauhar et al 2015] Jauhar, S. K., Dyer, C. and Hovy, E., 2015, Ontologically grounded multi-sense representation learning for semantic vector space models, *Proc. of NAACL-HLT*, 683-693. doi:10.3115/v1/N15-1070

[Kim 2006] Kim, Y. (2014). Convolutional neural networks for sentence classification. In *Proceedings of EMNLP*, 1746‑1751, Doha, Qatar.

[Klein et al 2006] Klein, G., Moon, B., Hoffman, R.F., 2006, Making sense of sensemaking I and II: *IEEE Intelligent Systems,* 21(4), 70–73. doi:10.1109/MIS.2006.75

[Landauer TK et al 1998] Landauer TK, Foltz PW, Laham D (1998) An introduction to latent semantic analysis. *Discourse Process* 25(2‑3):259‑284.

[Lebret et al 2013] Lebret, R., Collobert, R., 2013, Word emdeddings through hellinger pca. arXiv preprint arXiv:1312.5542, 2013.

[Le and Mikolov 2014] Le, Q., Mikolov, T., 2014, Distributed representations of sentences and documents., *Proc. Int'l Conf. on Machine Learning* 32(2), 1188-1196.

[Li et al 2018] Li, Y., Yang, T., 2018, Word embedding for understanding natural language: A survey. 83–104, 2018.

[Li and Jurafsky 2015a] Li, J., Jurafsky,D., 2015, Do multi-sense embeddings improve natural language understanding?, *Proc. of EMNLP*, 1722- 1732. doi:10.18653/v1/D15-1200.

[Maehara and Takenaka 2023] Maehara, T, Takenaka, Y., 2023, Generation of word embeddings for Japanese word sense disambiguate using paragraph embeddings in front and behind the target, in the 37th *Ann. Conf. JSAI.*

[Melamud et al 2016] Melamud, O., McClosky, D., Patwardhan, D., and Bansal, M., 2016, The Role of Context Types and Dimensionality in Learning Word Embeddings. *In Proc. Conf. North Am. Chap. of Assoc. Computational Linguistics: Human Language Technologies,* 1030–1040. doi:10.18653/v1/N16-1118

[Mignan 2011] Mignan, A., 2011, Retrospective on the Accelerating Seismic Release (ASR) hypothesis: Controversy and new horizons, *Tectonophysics* 2011, 505, 1–16. doi:10.1016/j.tecto.2011.03.010.

[Mignan 2014] Mignan, A., 2014, The debate on the prognostic value of earthquake foreshocks: A meta-analysis. *Sci. Rep.* 2014, 4, 4099. doi: 10.1038/srep04099

[Mikolov et al 2013a] Mikolov,T., Chen, K., Corrado, G., Dean, J., 2013, Efficient Estimation of Word Representations in Vector Space., doi: 10.48550/arXiv.1301.3781

[Mikolov et al 2013b] Mikolov, T., Sutskever, I., Chen, K., Corrado, G.S., Dean, J., 2013, Distributed representations of words and phrases and their compositionality. *Advances in Neural Information Processing Systems* 3111–3119, doi: 10.48550/arXiv.1310.4546

[Navigli et al 2013] Navigli, Roberto, David Jurgens, and Daniele Vannella. 2013. SemEval-2013 Task 12: Multilingual Word Sense Disambiguation. *In Proc.s of SemEval* 2013, pages 222–231.

[Nanjo et al 2006a] Nanjo, K.Z., Rundle, J.B., Holliday, J.R., Turcotte, D.L., 2006, Pattern informatics and its application for optimal forecasting of large earthquakes in Japan. *Pure Appl. Geophys*. 2006, 163, 2417–2432. doi:10.1007/s00024-006-0130-2

[Nanjo et al 2006b] Nanjo, K.Z., Holliday, J.R., Chen, C.C., Rundle, J.B., 2006, Turcotte, D.L. Application of a modified pattern informatics method to forecasting the locations of large future earthquakes in the central Japan. *Tectonophysics* 424, 351–366. doi: 10.1016/j.tecto.2006.03.043

[Neelakantan et al 2014] Neelakantan, A., Shankar, J., Passos, A. and McCallum, A., 2014, Efficient non-parametric estimation of multiple embeddings per word in vector space*, Proc. of EMNLP*, 1059-1069. Doi: 10.3115/v1/D14-1113.

[Ogata and Zhang 2006] Ogata, Y., 2006, Zhang, J. Space-time ETAS models and an improved extension. *Technophysics* 413, 13–23. doi: 10.1016/j.tecto.2005.10.016

[Ohsawa 2002] Ohsawa, Y., 2002, KeyGraph as Risk Explorer in Earthquake–Sequence, *J. Contingencies and Crisis Management* 10(3) 119-128 doi: 10.1111/1468-5973.00188.

[Ohsawa 2018] Ohsawa, Y., 2018, Regional Seismic Information Entropy to Detect Earthquake Activation Precursors, *Entropy* 20(11), 861; doi: 10.3390/e20110861

[Ohsawa et al 2022] Ohsawa, Y., Sekiguchi, K., Maekawa, T., Yamaguchi, H., Sun, HY., Kondo, S., 2022, Data Leaves as Scenario-oriented Metadata for Data Federative Innovation on Trust," *IEEE Int'l Conf. on Big Data*, 6159-6168, doi: 10.1109/BigData55660.2022.10020879.

[Ohsawa et al 2023] Ohsawa, Y., Kondo, S., Sun, Y., and Sekiguchi, K., 2023, Moving Direction Entropy as Index for Inter-community Activity, *Int'l Conf. on Knowledge-Based and Intelligent Information & Engineering Systems (KES), Proc. Comp. Sci.* 225, 4580–4587. doi: 10.1016/j.procs.2023.10.456

[Ohsawa 2024a] Ohsawa, Y., Maekawa, T., Yamaguchi, H., Yoshida, H., Sekiguchi, K., 2024, Collect and Connect Data Leaves to Feature Concepts: Interactive Graph Generation Toward Well-being, A*AAI Spring Symposium on the Impact of GenAI on Social and Individual Well-being*, https://arxiv.org/abs/2312.10375

[Ohsawa et al 2024b] Ohsawa, Y., Kondo, S., Sun, Y., Sekiguchi, K., 2024, Generating a Map of Well-being Regions using Multiscale Moving Direction Entropy on Mobile Sensors, *AAAI Spring Symposium on the Impact of GenAI on Social and Individual Well-being*, https://arxiv.org/abs/2312.02516

[Peters et al 2018] Peters, M.E., Neumann, M., Iyyer, M., Gardner, M., Clark, C., Lee, K., Zettlemoyer, L., 2018. Deep contextualized word representations. I*n Proc.Conf. North Am. Chap. Assoc. Computational Linguistics: Human Language Technologies* 1, 2227‑2237.

[Řehůřek 2022] Řehůřek, R., 2022, Gensim Word2vec embeddings https://radimrehurek.com/gensim/models/word2vec.html

[Reisinger et al 2010] Reisinger, J, Mooney, R.J., 2010, Multi-prototype vector-space models of word meaning. *In Proc. ACL,* 109‑117.

[Rothe and Schütze 2015] Rothe, S. and Schütze, H., 2015, AutoExtend: Extending Word Embeddings to Embeddings for Synsets and Lexemes. *In Proc. Ann. Meet.*

*Assoc. for Computational Linguistics and the Int'l Joint Conf. Natural Language Processing*, 1793–1803. doi:10.3115/v1/P15-1173 .

[Russell et al 2009] Russell, D. M., Pirolli, P., Furnas, G., Card, S. K., Stefik, M., 2009, Sensemaking workshop CHI 2009. *In CHI'09 Extended Abstracts on Human Factors in Computing Systems,* 4751–4754,

[Schütze et al 1993] Schütze, H., 1993, Word space. In Hanson, S.J., Cowan, J.D., Giles, C.L. (edr), *Advances in Neural Information Processing Systems* 5. 895‑902.

[Talman A et al 2018] Talman A, Yli-Jyra A, Tiedemann J (2018) Natural language inference with hierarchical Bilstm max pooling architecture. arXiv preprint arXiv:180808762

[Tian et al 2014] Tian, F., Dai, H., Bian, J., Gao, B., Zhang, R., Chen,E. and Liu, T.Y., 2014, A probabilistic model for learning multiprototype word embeddings, *Proc. of COLING*, 151-160.

[Tian et al 2024] Tian, W.X., Zhang, Y.X., Zhang, S.F., Zhang X T. 2024. Effect on the predictability of pattern informatics method related to selection of studied regions. *Acta Seismologica Sinica* 46（2）1−18. doi: 10.11939/jass.20220113

[Wang et al 2020] Wang, S., Zhou, W., Jiang, C.A., 2020, A survey of word embeddings based on deep learning. *Computing* 102, 717‑740. https://doi.org/10.1007/s00607-019-00768-7

[Weick 1993] Weick, KE., 1993, The Collapse of Sensemaking in Organizations: The Mann Gulch Disaster, *Administrative Science Quarterly* 38: 628-652. doi: 10.2307/2393339

[Wiedemann et al 2019] Wiedemann, G., Remus, S., Chawla, A., Biemann, C., 2019, Does BERT Make Any Sense? Interpretable Word Sense Disambiguation with Contextualized Embeddings. ArXiv, abs/1909.10430.

[Wyss and Habermann 1988] Wyss, M.; Habermann, R.E., 1988, Precursory seismic quiescence. *Pure Appl. Geophys*. 126, 319–332. doi: 10.1007/BF00879001

[Xu et al 2000] Xu, W, Rudnicky, A., 2000, Can artifcial neural networks learn language models? In Int'l Conf. Spoken Language Processing

[Zhang et al 2010] Zhang, Y., Jin, R., Zhou, ZH., 2010, Understanding bag-of-words model: a statistical framework. *Int J Mach Learn Cybern* 1(1–4):43–52

[Zhou et al 2019] Zhou, W., Ge, T., Xu, K., Wei, F., Zhou, M., 2019, BERT-based lexical substitution. *In Proc. Ann. Meet. Assoc. Computational Linguistics*, pages 3368–3373.